\documentclass[letterpaper, 10 pt, conference]{ieeeconf}  

\usepackage{textcomp}
\usepackage{amsfonts,amssymb} 
\usepackage{booktabs}
\usepackage{stfloats}
\usepackage{url}
\usepackage{verbatim}
\usepackage{graphicx}
\usepackage{cite}
\usepackage[version=4]{mhchem}
\usepackage{graphicx} 
\usepackage{epsfig} 
\usepackage{mathptmx} 
\usepackage{times} 
\usepackage{amsmath} 
\usepackage{amssymb}  
\usepackage[T1]{fontenc}
\usepackage{tabularx}  
\usepackage{hyperref}
\hypersetup{hidelinks,
	colorlinks=true,
	allcolors=black,
	pdfstartview=Fit,
	breaklinks=true}
\usepackage{multirow}
\usepackage{url}
\usepackage{makecell}
\usepackage{diagbox} 
\usepackage{bm}
\usepackage{graphicx} 
\usepackage{array}    
\usepackage[linesnumbered,ruled,vlined]{algorithm2e}

\IEEEoverridecommandlockouts                              
\title{\LARGE \textbf{VET: A Visual-Electronic Tactile System for Immersive Human-Machine Interaction}}

\author{
    Cong Zhang\textsuperscript{*1}, 
    Yisheng Yang\textsuperscript{*1}, 
    Shilong Mu\textsuperscript{*†1},  
    Chuqiao Lyu\textsuperscript{1},
    Shoujie Li\textsuperscript{1},
    Xinyue Chai\textsuperscript{1},  \\
    Wenbo Ding\textsuperscript{†1}
}

\begin{document}

\maketitle
\thispagestyle{empty}
\pagestyle{empty}

\begingroup
\renewcommand\thefootnote{}
\footnotetext{*These authors contributed equally to this work.}
\footnotetext{†Corresponding authors.}
\footnotetext{This work was supported by National Key R\&D Program of China (No.2024YFB3816000), Shenzhen Key Laboratory of Ubiquitous Data Enabling (No. ZDSYS20220527171406015), Guangdong Innovative and Entrepreneurial Research Team Program (2021ZT09L197), and Tsinghua Shenzhen International Graduate School-Shenzhen Pengrui Young Faculty Program of Shenzhen Pengrui Foundation (No. SZPR2023005).}
\footnotetext{\textsuperscript{1}6, Tsinghua University, Shenzhen, China, 518055.}

\endgroup

\begin{abstract}

In the pursuit of deeper immersion in human-machine interaction, achieving higher-dimensional tactile input and output on a single interface has become a key research focus. This study introduces the Visual-Electronic Tactile (VET) System, which builds upon vision-based tactile sensors (VBTS) and integrates electrical stimulation feedback to enable bidirectional tactile communication.  We propose and implement a system framework that seamlessly integrates an electrical stimulation film with VBTS using a screen-printing preparation process, eliminating interference from traditional methods. While VBTS captures multi-dimensional input through visuotactile signals, electrical stimulation feedback directly stimulates neural pathways, preventing interference with visuotactile information.  The potential of the VET system is demonstrated through experiments on finger electrical stimulation sensitivity zones, as well as applications in interactive gaming and robotic arm teleoperation. This system paves the way for new advancements in bidirectional tactile interaction and its broader applications.

\end{abstract}

\section{Introduction}

With the continuous advancement of input devices, user demands have expanded beyond simple actions such as tapping or pressing. Increasingly, users require more detailed input control, including force and directional precision. This demand is particularly evident in fields such as robotic teleoperation and virtual reality\cite{adami2021effectiveness}. Traditional input devices are based on handles\cite{naceri2021vicarios}, vision\cite{meng2023virtual}, and wearable devices\cite{pyun2022materials} to achieve natural interaction instructions and have been widely used in various fields. At the same time, VBTS have great potential in robot operation and interaction due to their simple manufacturing and rich information acquisition. However, devices with only one input module are often not enough to complete all operation tasks well. The integration of VBTS with haptic feedback output represents a significant step forward, driving overall progress and innovation in input device technology.

\begin{figure}[htbp]
    \centering
    \includegraphics[width=0.45\textwidth]{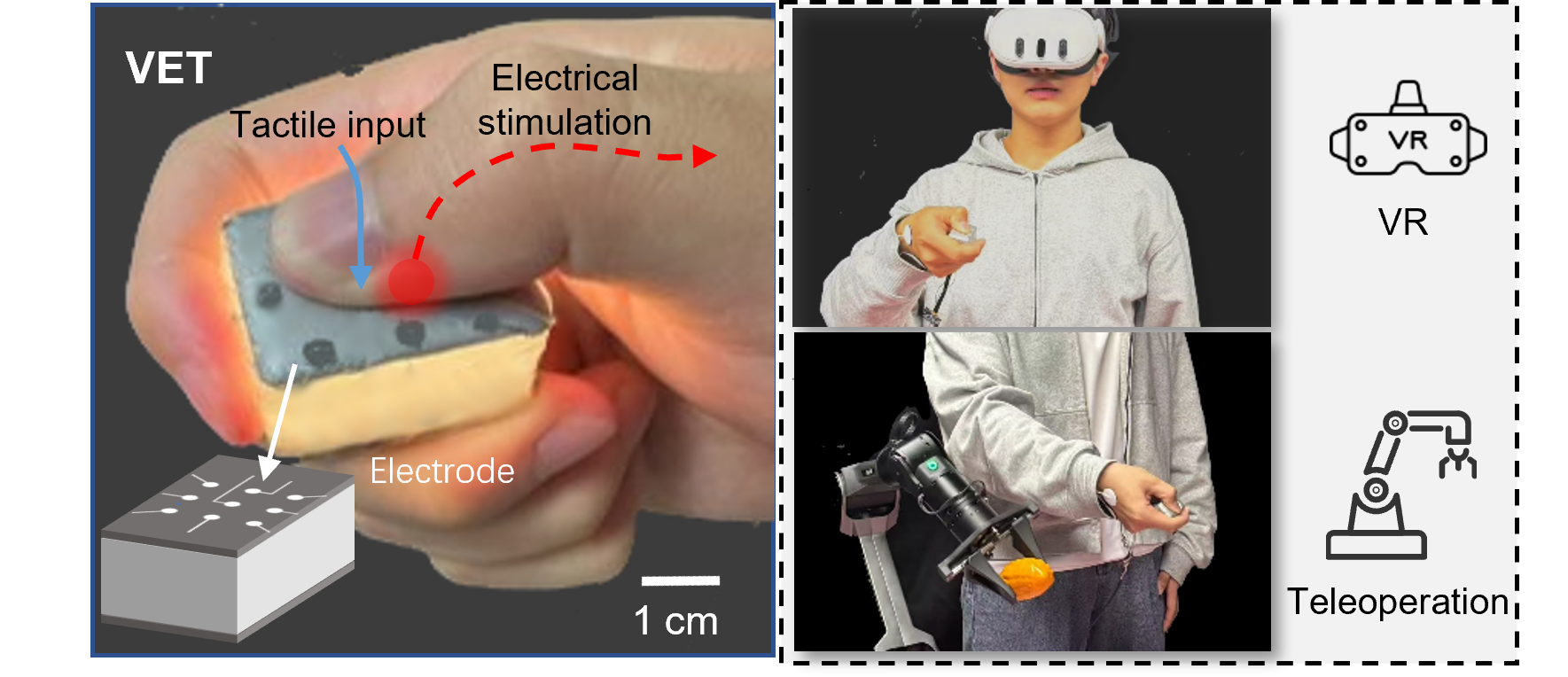}
    \caption{Illustration of the most critical components of the VET system. VET achieves both tactile sensing and haptic feedback on the same surface, which can be applied to multiple fields, such as virtual reality and teleoperation of robotic arms.}
    \label{fig1}
\end{figure}

Currently, haptic feedback is primarily achieved through mechanical force\cite{pyun2022materials} and vibration mechanisms\cite{yu2019skin}. However, integrating these two feedback modalities into a VBTS can significantly impact its sensing performance. Force feedback introduces physical disturbances to the contact state between the sensor and the object, while vibration feedback may induce sensor oscillations, thereby reducing the accuracy of data acquisition. These interference issues hinder the formation of effective closed-loop control between input and feedback, severely limiting system performance in high-precision applications.

To address this challenge, this study proposes a system that employs electrical stimulation for haptic feedback. Using electrical stimulation, the system provides precise and diverse feedback modalities, including information prompts, and intensity alerts, while mitigating the adverse effects of traditional haptic feedback on the sensor performance.

The main contributions of this work are as follows:
\begin{itemize}
\item We propose and implement a system framework that combines electrical stimulation film with VBTS by combining the screen printing preparation process to cleverly integrate electrical stimulation film on the surface of the vision-based tactile sensor, eliminating the interference of traditional methods.

\item Through experimental validation in both interactive applications and teleoperation scenarios, we substantiate the advantages of the proposed system, offering innovative perspectives for the development of high-accuracy human-robot interaction systems.
\end{itemize}

The remainder of this paper is organized as follows: Section II reviews related work, Section III details the system design and implementation, Section IV presents the experimental methods and results, and Section V concludes the paper and discusses future research directions.

\begin{figure*}[thpb]
  \centering
  \includegraphics[width=1.0\linewidth]{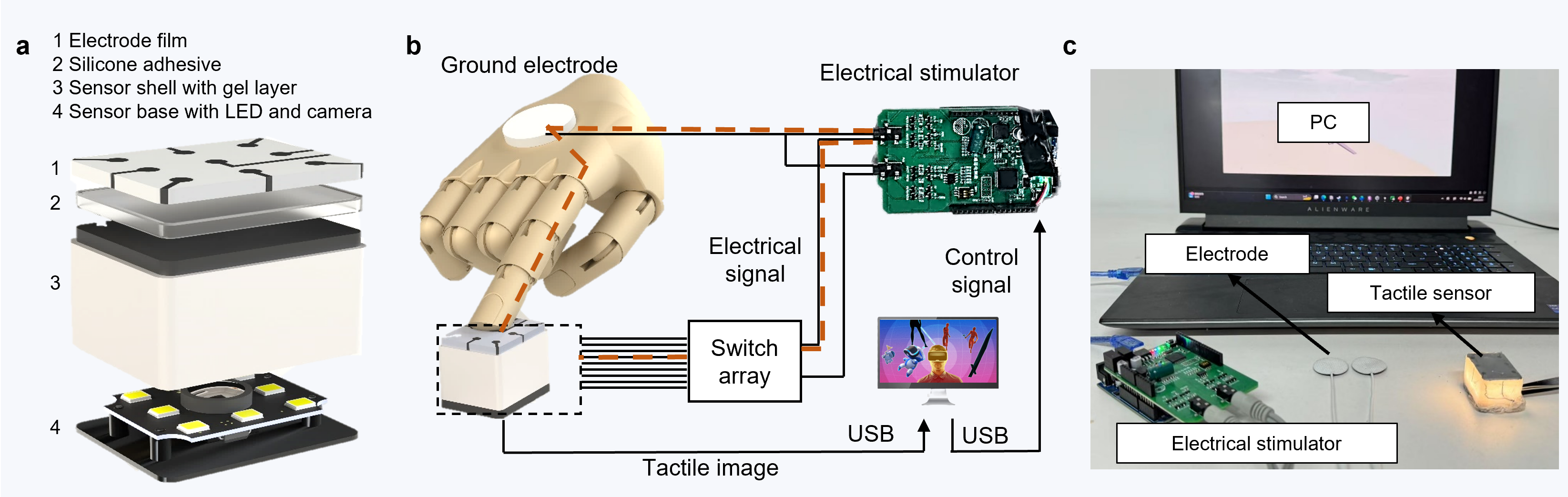}\\
  \caption{Framework diagram of bidirectional tactile human-machine interaction. (a) Internal structure diagram of VBTS. (b) Components of VET and an introduction to their connection methods. (c) Physical demonstration diagram of the VET system. }
  \label{fig2}
\end{figure*}

\section{Related work}
\subsection{Human-machine interface} Lin \textit{et al.} \cite{lin2022super} developed a super-resolution wearable electrotactile system achieving 76 dots/cm² resolution for VR texture simulation. In robotics, Rahimi \textit{et al.} \cite{rahimi2019dynamic} integrated fingertip-based electro-tactile displays with dynamic spatiotemporal patterns for enhanced tactile feedback. For prosthetics, Galofaro \textit{et al.} \cite{galofaro2022rendering} validated neuromuscular electrical stimulation for immersive haptic feedback, overcoming the challenges of traditional tactile systems. Hybrid approaches (e.g., FinGAR \cite{yem2017wearable}) combining electrical and mechanical stimuli further improve realism but require synchronization. Current limitations include electrode durability and user-specific sensitivity \cite{mohammadzadeh2024nafion}.
\subsection{Vision-based tactile sensor}
VBTS, as innovative optical sensors, have been widely applied in robotic perception. Lin \textit{et al.}\cite{lin2023dtact} introduced a marker-free design for 6D force estimation, but its reliance on external illumination limited its practicality. In contrast, 9DTact \cite{10342722 } builds on these advancements by integrating high-precision 3D shape reconstruction and generalizable 6D force estimation into a single, compact, and easily manufacturable sensor.

\subsection{Haptic feedback}
Researchers have been actively pursuing a feedback architecture that minimizes interference with VBTS data acquisition from conventional feedback mechanisms. Electrical stimulation not only reduces space occupancy but also minimizes its impact on visual-tactile sensors. Recent advancements in electrotactile technology have improved haptic feedback and texture simulation. Lin \textit{et al.} \cite{lin2022super} developed a super-resolution wearable system. Galofaro \textit{et al.}
\cite{galofaro2022rendering} introduced neuromuscular electrical stimulation for immersive haptic feedback in virtual environments. Guo \textit{et al.} \cite{guo2022skin} developed on-skin stimulation devices for realistic feedback in human-machine interfaces. Huang \textit{et al.} \cite{huang2022recent} reviewed multi-modal haptic technologies combining electrical and other stimulation methods to enhance wearable interfaces.

While previous studies have explored VBTS and electrical stimulation separately, a seamless integration of these technologies for bidirectional tactile interaction remains an open challenge. Our proposed VET System bridges this gap by integrating electrical stimulation feedback directly onto VBTS surfaces using a screen-printing process, eliminating the interference issues of traditional methods. 
\begin{figure}[htbp]
    \centering
    \includegraphics[width=0.45\textwidth]{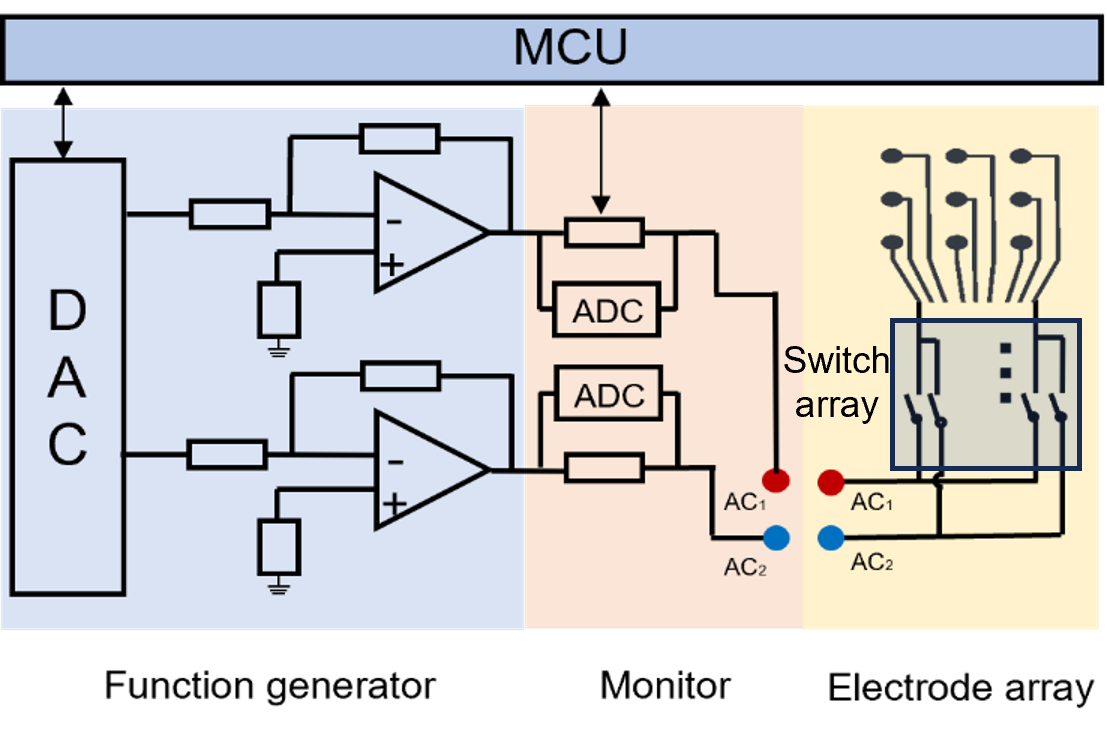}
    \caption{Schematic diagram of the electrical control system, which consists of three parts. The function generator generates the desired stimulation current AC1 and AC2, the current monitor ensures the consistency of output perception by feedback control, and the switch array manages the state of each electrode separately. ADC, analog-to-digital converter. }
    \label{fig3}
\end{figure}
\section{Methods and Design}
\subsection{System framework overview}

The proposed tactile interaction system integrates sensing and feedback mechanisms, consisting of an electrical stimulation film, silicone elastomer layer, electrical stimulation generator, and terminal unit. The system is compact (20 × 30 × 20 mm), portable, and features a computer interface for easy installation.

Upon fingertip contact, the sensor transmits tactile data (such as contact force, contour, and motion trajectory) to the terminal, while electrical stimulation provides haptic feedback. The operator can modulate the current’s frequency, intensity, and direction to produce varied feedback (Fig. 2b).

This system combines high-fidelity tactile perception with programmable feedback, enabling efficient bidirectional information transfer and significantly enhancing human-machine interaction and immersion.

\subsection{Working principle}

To ensure stable and effective electrical stimulation, VET uses a programmable dual-channel stimulator. The module, with independent channels, enables precise control of stimulation parameters. Its internal architecture (Fig. 3) includes a function generator, monitor, and electrode array. The function generator creates the stimulation waveform, which is monitored and regulated via ADCs before delivery to the electrode array. The module has a static power consumption of 130 mA, with a startup current of 250 mA, and can be powered through a USB interface for easy integration. The stimulation waveform is optimized for stability and precision, ensuring consistent effects. Based on Fig. 4a, the electrostimulation targets the Meissner corpuscles and Merkel cells\cite{biswas2021haptic}, sensitive in the 0.5 Hz to 100 Hz range, with 50 Hz used for experiments unless specified otherwise.


Various grounding configurations exist, including back\cite{yem2018effect}, ring\cite{yoshimoto2015material}, palm\cite{rahimi2019dynamic}, dorsum\cite{tezuka2016presentation}, coaxial\cite{kaczmarek2017interaction} , and coplanar\cite{lin2024tactex} placement. Among these, back, ring, palm, and dorsum are most commonly used in wearable devices. The key difference lies in the location of the grounding electrode, which impacts stimulation distribution. Table 1 summarizes the advantages and disadvantages of these configurations. To allow users to operate seamlessly with different fingers without interference, this study adopts the back grounding electrode configuration.  

 \begin{figure*}[htbp]
    \centering
    \includegraphics[width=1\textwidth]{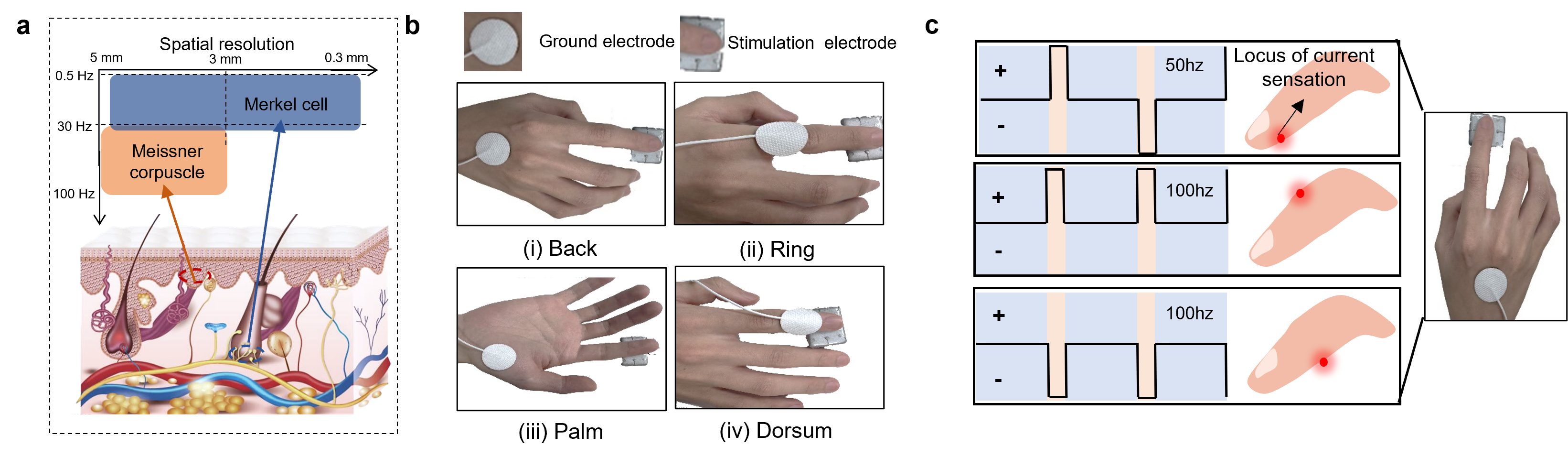}
    \caption{The principles, methods, and effects of electrical stimulation. (a) Sensory corpuscles involved in skin electrical stimulation and their response frequency and spatial resolution. (b) Picture of the grounding electrode placement position. (c) When the grounding electrode is placed on the back, the perceived locations on the hand are under electrical stimulation currents of different polarities (positive and negative).  }
    \label{fig4}
\end{figure*}

\begin{table}[h!]
  \centering
  \renewcommand{\arraystretch}{1.2} 
  \caption{Comparison of the grounding electrode placement methods.}
  \label{tab:control-comparison}
  \setlength{\tabcolsep}{6pt} 
  \resizebox{0.45\textwidth}{!}{ 

  \begin{tabular}{l|p{3.5cm} p{3.5cm}}  
    \toprule
    \textbf{Type} & \textbf{Advantages} & \textbf{Disadvantages} \\
    \midrule
    Back & 5 fingers share one grounding electrode.  
         & Stimulation happens on both the inner finger and palm. \\

    Ring & Requires small current.  
         & 5 grounding electrodes. \\

    Palm & 5 fingers share one grounding electrode.  
         & Inconvenient for hand manipulation. \\

    Dorsal & Stimulation is focused on the fingers being stimulated.  
           & 5 grounding electrodes. \\
    \bottomrule
  \end{tabular}
  }

\end{table}

The spatial distribution of the electrical stimulation is influenced by the polarity of the applied signals, as demonstrated in Fig. 4c. When the stimulation consists solely of positive pulses, the perceived sensation is more pronounced at the upper region of the fingertip. Conversely, when only negative pulses are applied, the sensation shifts toward the lower region of the fingertip. However, when alternating positive and negative pulses are used, the primary perception of stimulation is concentrated precisely at the contact point between the finger and the electrode, ensuring targeted and effective neuromodulation.

\begin{figure}[htbp]
    \centering
    \includegraphics[width=0.45\textwidth]{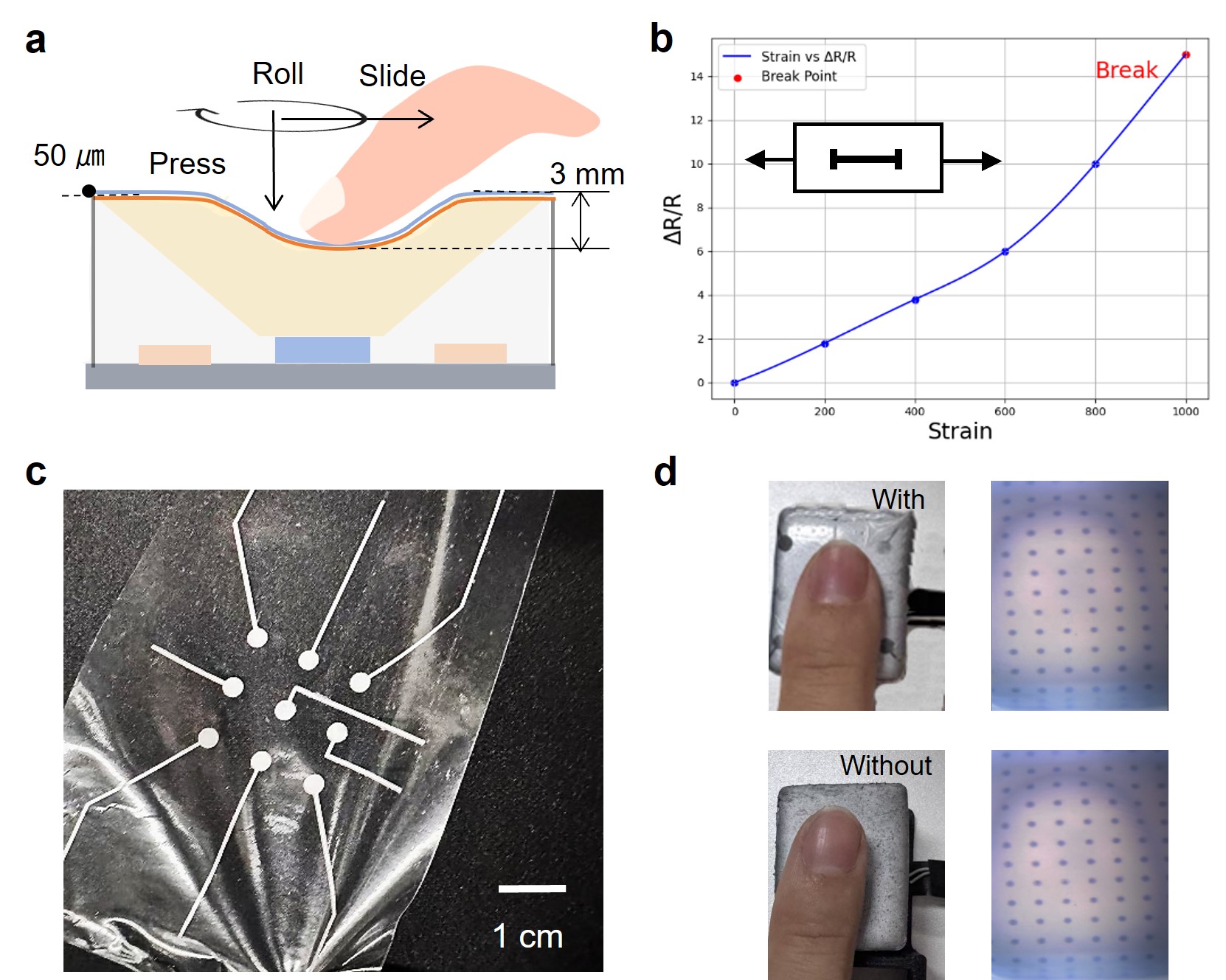}
    \caption{Principle of electrode stimulation film. (a) Schematic diagram of finger pressing on VET. (b) Verification of film deformation capabilities. (c) Physical representation of the film after screen printing. (d) Comparison of internal visuotactile images of VET with and without electrical stimulation film. }
    \label{fig5}
\end{figure}

The system utilizes a VBTS, as shown in Fig. 2a. Contact with the semi-transparent gel surface induces localized thinning (reducing light reflection) and peripheral thickening (enhancing reflection). A camera detects these intensity variations, enabling 3D shape reconstruction via algorithms. This real-time force feedback facilitates closed-loop adjustment of electrical stimulation parameters, ensuring targeted neuromodulation aligned with tactile interactions.

In Fig. 5a, the experiment reveals that when the sensor is pressed, the lower limit of the sensor's surface is approximately 30 mm, with a deformation rate of about 300\%. The tensile test of the electrostimulation film, shown in Fig. 5b, demonstrates the structural stability of the material. To verify that the film does not interfere with the tactile sensing performance of the VBTS, we compared the tactile images both with and without the film. The results show that the tactile images remain largely consistent in both cases, indicating the film does not impact VBTS sensing.

\subsection{VET fabrication}

This section outlines the fabrication of the flexible electronic film, including material selection, screen printing, laser perforation, backside printing, adhesion, and touch testing. The process ensures the integration of a conductive and stretchable layer with stable adhesion to the VBTS (Fig. 6).

\begin{figure*}[htpb]
  \centering
  \includegraphics[width=1\textwidth]{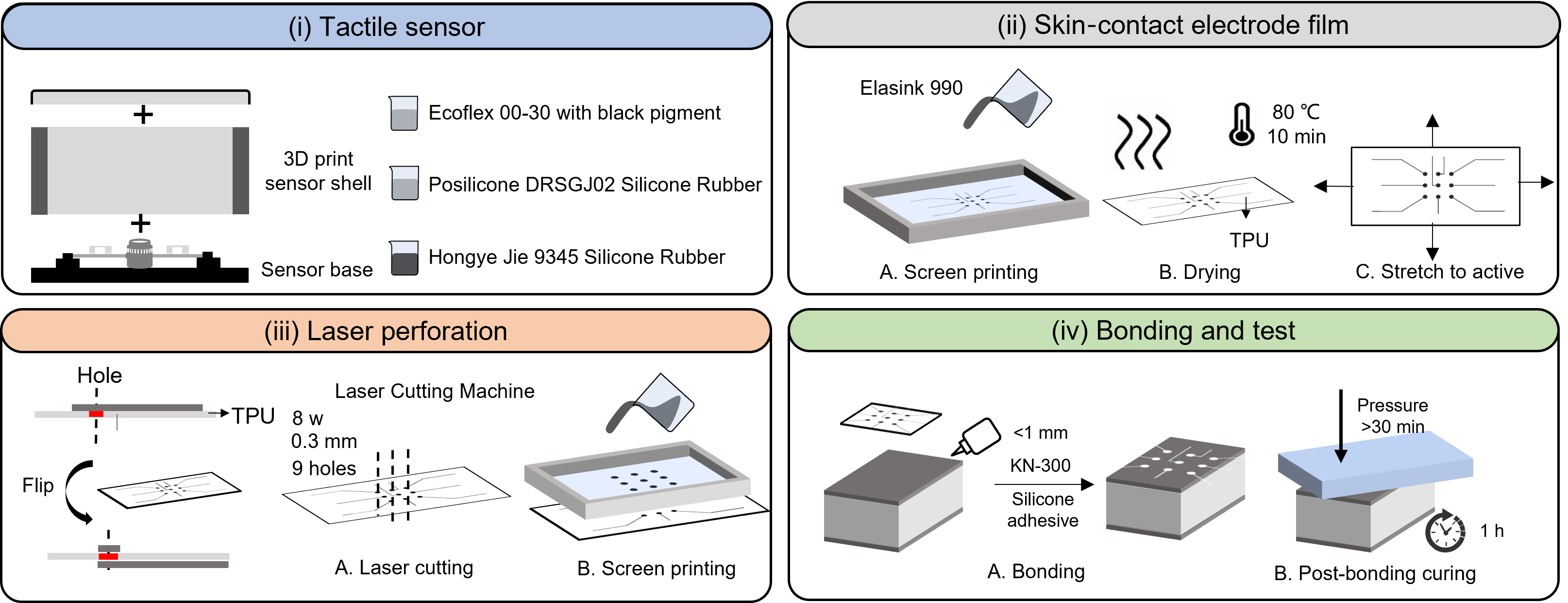}
  \caption{Fabrication process of the entire device. (I) Fabrication process of VBTS. (ii) Screen printing of the electrical stimulation film with the complete circuit. (iii) Laser cutting machine perforating at the electrode positions followed by screen printing the electrodes on the reverse side. (iv) Bonding the electrical stimulation film to VBTS and conducting functional testing.}
  \label{fig6}
\end{figure*}

\textbf{Step 1: Fabrication of VBTS.}  
The fabrication process involves preparing the substrate, depositing thin films (e.g., spin-coating/chemical vapor deposition) to create functional layers, using light and etching to pattern the electrodes, and adding protective layers.

\textbf{Step 2: Screen Printing of Circuits.}  
A polymer mesh screen was used for ink deposition, following these steps \cite{tang2021multilayered}:
1) Ink Preparation: The conductive ink was stirred for 10 minutes to ensure uniformity.
 2) Screen Printing: A polymer mesh screen with a 200-mesh count was used.
3) Drying Process: The printed ink was dried at 80°C for 10 minutes, or at room temperature for 30 minutes until the color changed from gray to off-white.
4) Stretch Activation: The ink was mechanically stretched in two perpendicular directions (50\% strain) to activate conductivity.

\textbf{Step 3: Laser Cutting.}  
Laser cutting was used to perforate the electrode areas of the film with a diameter of 0.3mm\cite{xin2021process}. After flipping the film, a second screen printing process was performed on the electrode areas to ensure precise electrical contact.

\textbf{Step 4: Adhesion.}  
The printed film was bonded using silicone adhesive and pressed with a heavy object for 30 minutes. Afterward, it was left to rest for one hour to ensure proper adhesion.

\section{Experiment}
The following experiments evaluate the effectiveness VET in enhancing interaction and control. The first experiment examines the spatial sensitivity of electrotactile feedback on different finger regions. The second explores VET's immersive capabilities in a flight simulation game and robotic arm teleoperation for improved grasping precision. These studies highlight VET’s potential in both entertainment and robotic applications.

\subsection{Spatial electrotactile sensitivity mapping in finger zones}

\begin{figure*}[thpb]
  \centering
  \includegraphics[width=1.0\linewidth]{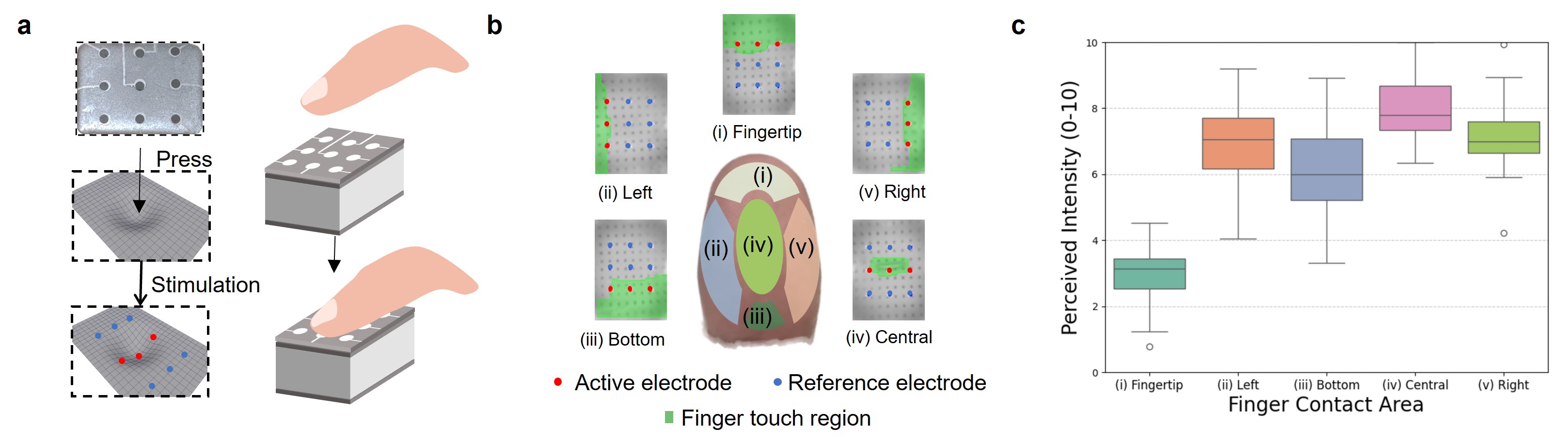}
  \caption{Experiment investigating the sensitivity of different finger regions to electrical stimulation. (a) Experimental principle flowchart: When the finger presses on VET, it detects the finger pressing area and applies corresponding electrical stimulation to that region. (b) Five designated areas on the finger which are capable of making contact with the electrode. (c) Box plot of experimental results testing the sensitivity of different hand positions to electrical stimulation.
}\label{fig7}
\end{figure*}

We conducted an experiment to investigate the spatial dependency of electrotactile perception across different finger subregions using the VET system. As shown in Fig. 7a, the experimental procedure involved applying electrical stimulation corresponding to specific pressure conditions while participants interacted with the sensor. The finger was divided into five anatomical zones: fingertip, left, bottom, ventral, and tight (Fig. 7b). Electrical stimulation was applied to the center of each zone to assess variations in tactile perception across these regions.

Participants performed trials under two conditions: baseline (sensor-only pressure recording) and stimulated (electrical pulses). Each trial consisted of a 2-second contact phase followed by a 10-second rest interval to minimize sensory adaptation. The pressure profiles were segmented by zone, and the statistical analysis of the perceived intensity revealed significant spatial sensitivity variations, as illustrated in Fig. 7c. The fingertip and central regions exhibited the highest sensitivity scores, attributed to a higher density of mechanoreceptors and uniform pressure distribution. In contrast, the left and right zones showed lower sensitivity due to fewer nerve endings and uneven skin curvature.

\subsection{VET-based bidirectional tactile interaction application
}

\begin{figure*}[thpb]
  \centering
  \includegraphics[width=1.0\linewidth]{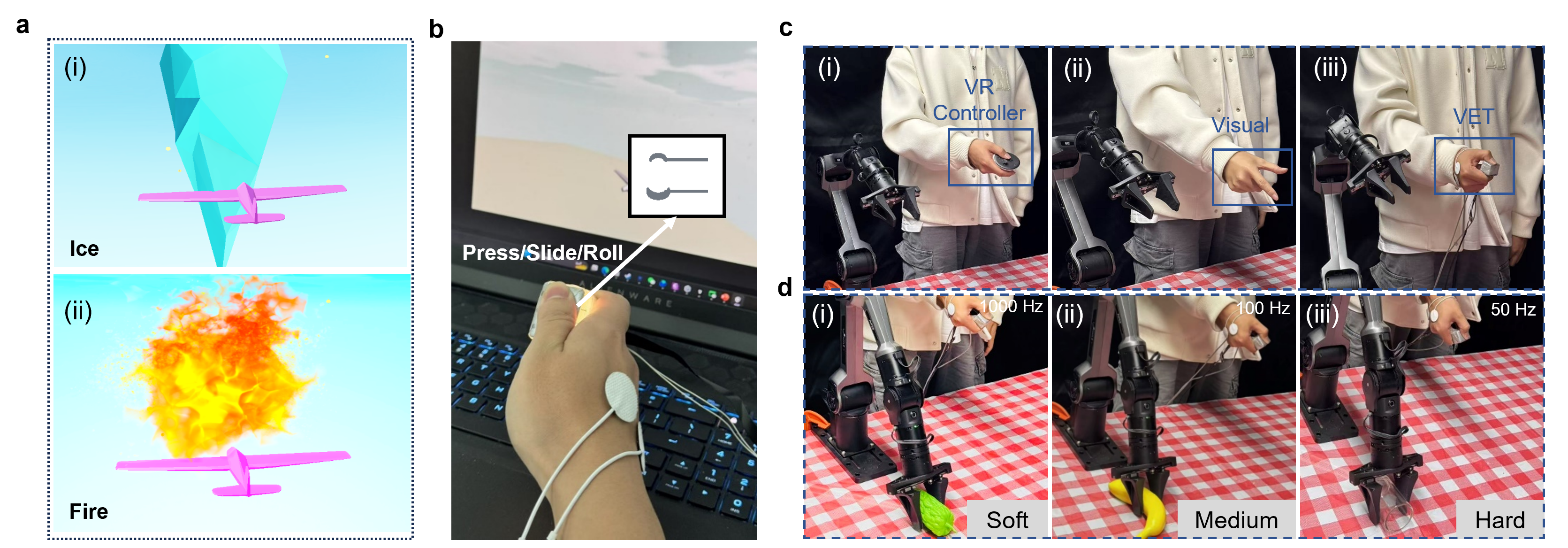}
  \caption{Demonstration of the VET System in Specific Application Scenarios. (a) Electrotactile feedback in the "ice" and "fire" zone. (b)  Unity-based 3D flight control environment with real-time haptic feedback integration. (c) Different control approaches for robotic arms. (d) Electronic feedback when robotic arms grasp objects of different softness and hardness.
}\label{fig8}
\end{figure*}
We evaluated the VET system’s immersive potential using a Unity-based flight simulation where pilots navigate through dynamically generated ice and fire zones while collecting cargo \cite{lee2020mimicking,handler2021mechanosensory}. The system employs a unified tactile interface (integrating a miniaturized visuotactile sensor with an electrotactile array) to enable bidirectional, single-finger interactions. Optical flow analysis maps directional pressures to flight dynamics, while context-sensitive electrotactile feedback signals environmental hazards: low-frequency (10 Hz, low current) vibrations simulate the slipperiness of ice, and high-frequency (50 Hz, high current) pulses warn of fire zones. Additionally, transient high-intensity pulses upon collisions and rhythmic feedback during sustained contact further enhance situational awareness. Users reported a significantly enhanced sense of immersion and increased gameplay enjoyment when using the VET system. This multimodal approach improves environmental discriminability and control precision, underscoring the promise of tactile systems in advanced human-machine interfaces.

Within entertainment contexts, the VET system demonstrates significant efficacy by delivering immersive, physically consistent feedback that enhances user control and situational awareness. This approach is equally effective in robotic arm teleoperation, where it contributes to precise, real-time adjustments and intuitive interaction. To better understand the comparative advantages of these systems, Fig. 8c visualizes the three control methods— VR Controller, Vision-based, and VET. The detailed comparison in Table \ref{tab:control-comparison} further highlights their respective strengths, such as precision and safety, across dynamic and complex tasks.

\begin{table}[h!]
  \centering
  \renewcommand{\arraystretch}{1} 
    \caption{Comparison of Control Systems}
  \label{tab:control-comparison}
  \setlength{\tabcolsep}{6pt} 
  \resizebox{0.45\textwidth}{!}{ 
  
  \begin{tabular}{l|ccc}  
    \toprule
    \textbf{Feature} & \textbf{VR Controller} & \textbf{Vision} & \textbf{VET} \\ 
    \midrule
    Perception dimension  & Two-dimensional & Visual & Multi-dimensional \\
    Feedback    & Vibration  & none & Electronic stimulation \\ 
    Robustness  & High & Low & High \\
    \bottomrule
  \end{tabular}
  }

\end{table}

To demonstrate the advantages of the VET system in specific grasping tasks, we designed a scenario where the robotic arm encounters objects with varying levels of softness and hardness. In this scenario, the VET system adjusts the frequency of electrical stimulation to provide the operator with distinct tactile sensations, allowing for more precise control over the grasping force. This feedback enables the operator to fine-tune the grip strength in real time, thereby enhancing the accuracy and effectiveness of the grasping process.

\section{Discussion and Future Work}

This study presents the VET system as an innovative solution for bidirectional tactile interaction in human-machine interfaces. By integrating electrical stimulation feedback with vision-based tactile sensing through a screen-printed electrode film, the system achieves three critical advancements: 1) Elimination of interference between sensing and feedback modalities through spatial decoupling of visuotactile acquisition and neural stimulation pathways; 2) Development of a scalable fabrication process enabling seamless sensor-actuator integration within a compact 20 × 30 × 20 mm footprint; 3) Demonstration of enhanced interaction fidelity in both virtual environments and robotic teleoperation. This work establishes a foundation for closed-loop tactile interaction systems that combine high-resolution environmental sensing with precise haptic feedback capabilities.

Future research should focus on improving the system’s adaptability and accessibility. This includes incorporating multidimensional tactile sensing to expand user input capabilities and developing adaptive algorithms to personalize electrical stimulation for diverse user needs. Besides, exploring the VET system’s applications in robotics could open new possibilities for human-machine collaboration, while advancements in hardware miniaturization and wireless connectivity may enhance portability for real-world deployment. 

\bibliographystyle{ieeetr}
\bibliography{refe}
\end{document}